\documentclass{style}%

\usepackage{graphicx, xcolor, tcolorbox}
\usepackage{booktabs, array}
\usepackage[multiple]{footmisc}
\usepackage{float}
\usepackage{cleveref}

\usepackage{todonotes}

\title[]{A comparative study of artificial intelligence and human doctors for the purpose of triage and diagnosis
}

\author[Razzaki et al.]{{
Salman Razzaki$^*$, Adam Baker$^*$, Yura Perov$^*$, Katherine Middleton$^*$, Janie Baxter$^*$, Daniel Mullarkey$^*$, Davinder Sangar$^*$, Michael Taliercio$^*$, Mobasher Butt$^*$, Azeem Majeed$\textsuperscript{\textdagger}$, Arnold DoRosario$\textsuperscript{\textdaggerdbl}$, Megan Mahoney$\textsuperscript{\textsection}$ and Saurabh Johri$\textsuperscript{*,\textparagraph}$
}
\vspace{2.5mm}
\affil{\hspace{0.005em}$\textsuperscript{*}$ Babylon Health.}%
\affil{$\textsuperscript{\textdagger}$ School of Public Health, Faculty of Medicine, Imperial College London.}%
\affil{$\textsuperscript{\textdaggerdbl}$ Northeast Medical Group, Yale New Haven Health.}%
\affil{$\textsuperscript{\textsection}$ Division of Primary Care and Population Health, School of Medicine, Stanford University.}%
\affil{\hspace{-0.005em}$\textsuperscript{\textparagraph}$ Corresponding author: Saurabh Johri, saurabh.johri@babylonhealth.com.}
\vspace{2em}
}%

\usepackage[hyperfootnotes=false]{hyperref} 
\hypersetup{colorlinks,citecolor=blue,linkcolor=blue,urlcolor=blue}

\hypersetup{draft}

\begin{document}

\begin{frontmatter}
\maketitle

\begin{abstract}
Online symptom checkers have significant potential to improve patient care, however their reliability and accuracy remain variable. We hypothesised that an artificial intelligence (AI) powered triage and diagnostic system would compare favourably with human doctors with respect to triage and diagnostic\footnotemark{} accuracy. We performed a prospective validation study of the accuracy and safety of an AI powered triage and diagnostic system.  Identical cases were evaluated by both an AI system and human doctors. Differential diagnoses and triage outcomes were evaluated by an independent judge, who was blinded from knowing the source (AI system or human doctor) of the outcomes. Independently of these cases, vignettes from publicly available resources were also assessed to provide a benchmark to previous studies and the diagnostic component of the MRCGP\footnotemark{} exam. Overall we found that the Babylon AI powered Triage and Diagnostic System was able to identify the condition modelled by a clinical vignette with accuracy comparable to human doctors (in terms of precision and recall). In addition, we found that the triage advice recommended by the AI System was, on average, safer than that of human doctors, when compared to the ranges of acceptable triage provided by independent expert judges, with only a minimal reduction in appropriateness.
\vspace{1em}
\end{abstract}
\end{frontmatter}

\section{INTRODUCTION}
\label{sec:intro}
\addtocounter{footnote}{-2} 
\stepcounter{footnote}\footnotetext{The term ``diagnosis'' is used for ease of reference, as shorthand to suggest the matching of symptoms with diseases and/or conditions. However, we are not suggesting that online symptom checkers are diagnosing users, or that the Babylon Chatbot provides a diagnosis.}
\stepcounter{footnote}\footnotetext{Membership of the Royal College of General Practitioners.}

Online symptom checkers are a convenient and valuable resource for users to better understand the underlying cause(s) of their symptoms and to receive advice on the most appropriate point of care. Typically, symptom checkers cater to three healthcare needs of a patient. First is the provision of information, wherein a patient may seek to know more about the symptoms or conditions that they know or think they have. Secondly, a patient may want to know whether their symptoms require treatment or further investigation; this is medical triage and involves directing patients to the most suitable location within an appropriate time frame.  The appropriate action depends on the nature and urgency of the symptoms or their underlying cause, which might require further investigation. Finally, patients may want to understand the conditions that might be responsible for their symptoms. This corresponds to diagnosis or ``differential diagnosis'' and is typically performed by an experienced medical practitioner. 

Symptom checkers have the potential to alleviate the pressure on overly burdened healthcare systems. For this to happen, healthcare professionals and the wider public must have confidence in the performance of symptom checkers and applications of AI to medicine more generally. Previous work has investigated the diagnostic and triage accuracy of competing symptom checkers and highlighted significant variation in terms of clinical accuracy \citep{semigran2015evaluation}. Whilst providing a useful benchmark, that study did not assess the accuracy of symptom checkers against the gold-standard performance of human doctors. This was assessed in a follow-up study, where the authors noted that doctors significantly outperform symptom checkers, providing a valuable contribution to our understanding of comparative diagnostic performance \citep{semigran2016comparison}. However, the method used in this follow-up study did not adequately assess the information gathering process through which patients typically interact with symptom checkers or doctors, and so the conclusions are not based on a fair or realistic comparison. Diagnostic accuracy is not routinely measured in clinical practice, but a wide range of studies have attempted to estimate the incidence of diagnostic error. Irrespective of whether the true error rate lies closer to the 10-20\% found in autopsy studies \citep{graber2013incidence} or the 44\% found in a study analysing the correlation of diagnostic accuracy with doctor confidence \citep{meyer2013physicians}, it is critical to perform a fair assessment of how a doctor takes a history and establishes a diagnosis when comparing against symptom checkers.

In this study we adopt a semi-naturalistic, role-play paradigm that simulates a realistic consultation between a patient and either our Triage and Diagnostic System or human doctor. Based on the assessment technique used throughout medical school and post-graduate medical qualifications (Objective Structured Clinical Examinations [OSCE]), this protocol was designed to assess not only the clinical (diagnostic and triage) accuracy, but also the ability to gather all of the relevant information from the patient i.e. to take a history.

\section{The Babylon Triage and Diagnostic System}
\label{sec:babylon_td_system}

The Babylon Triage and Diagnostic System\footnote{
As of the current date (June 2018), the model of Babylon Triage and Diagnostic System being evaluated in the current study is not released yet.
}\footnote{
It should be noted that, this paper is for general information and academic purposes, and to analyse the use of online symptom checkers in healthcare. The Babylon Triage and Diagnostic System is referred to in this paper to facilitate discussion on this topic, but this paper is not designed to be relied upon for any other purpose.
} is designed to provide users with triage advice alongside an explanation of why this action has been suggested; this consists of any reported symptoms that require urgent attention, and/or a list of possible causes for the user's symptoms. A comprehensive description of the system that powers the Babylon Triage and Diagnostic System is outside of the scope of this paper, however we provide a brief summary of this system by way of background.

The Babylon Triage and Diagnostic System -- a new implementation after the previous generation \citep{middleton2016sorting} -- is based on a Probabilistic Graphical Model (PGM) \citep{koller2009probabilistic} of primary care medicine, which models the prior probabilities of diseases and the conditional dependencies between diseases, symptoms and risk factors via a directed acyclic graph. The structure of the graph (i.e., the connections between diseases, symptoms and risk factors) is created by medical experts and reviewed from a modelling perspective. The prior probabilities of diseases and risk factors are obtained from epidemiological data, where available. Conditional probabilities (for example, the probability of a symptom occurring given a disease) are obtained through elicitation from multiple independent medical experts. 

Once constructed and parameterised, the model is used to reason about the possible underlying diseases that cause the user-entered symptoms and risk factors, using novel Bayesian inference methods \citep{douglas2017universal,cheng2000ais,wainwright2008graphical,gu2015neural}. This allows the AI powered Triage and Diagnostic System to output the most likely causes of the symptoms entered by a user, and also generate follow up questions that provide the most information to confirm or rule out the disease under consideration.

The triage capability is based on an extension of the PGM with a utility model \citep{koller2009probabilistic}. This combines the likelihood of each disease with the potential harm caused by that disease, under all possible triage decisions. The triage decision that is recommended is the one that minimises the expected harm to the patient, while also penalising overtriaging. To guarantee the safe triage of patients with symptoms that require urgent treatment (regardless of their underlying cause), the utility model is augmented with a set of rules that dictate a specific triage action where a particular combination of symptoms (so-called ``red-flag'' symptoms) are present. The system is designed to identify one of six triage actions: ``call an ambulance'', ``go to A\&E/ER'', ``urgent GP'' (i.e., within 6 hours), ``non-urgent GP'' (i.e. within a week), ``pharmacy'' and ``self-care''.

\section{Iterative validation and development}
\label{sec:val_and_development}

During development, the performance of the model is continuously assessed through an internal validation process to identify areas of weakness that could be improved. Validation is performed against a set of simulated clinical vignettes that are created by medical experts within an internal web tool. Each clinical vignette is written to reflect a realistic presentation of a patient with a particular disease or condition, containing the patient's symptoms, past medical history and basic demographic information such as age and sex. The author of the vignette is instructed to also enter any absent symptoms or risk factors that a doctor would be likely to enquire about during a typical consultation. Where possible, symptoms and risk factors match those in the model to allow the system to recognise these entities automatically. However, to avoid the bias of only including those entities that are present in the model, the author of the clinical vignette is allowed to enter any other symptoms or risk factors via free text. 


All clinical vignettes are assessed by our model in two modes: 1) by only providing to the model those symptoms that are elicited by the chatbot, and 2) by providing to the model all symptom and risk-factor entities listed on the vignette. This helps evaluate not only the accuracy of the model, but also the history-taking ability of the system; because the chatbot must choose the appropriate questions to ask it is not guaranteed that all entities that exist on the vignette will be available to the model.

The clinical vignettes are also assessed by doctors. Each doctor independently reviews the vignette and enters their own differential diagnosis and triage outcomes. This allows the predictions from the model (the list of possible causes and recommended triage action) to be compared not only against the disease modelled by the clinical vignette but also against the consensus of multiple doctors. 

Those vignettes against which the model performs poorly are analysed in depth by a combination of doctors and scientists to identify parts of the model that require review, and this process of validation and iterative improvement results in continuous improvement of the model. As is standard practice in machine learning, any vignettes used to train or otherwise inform the structure or parameters of the model are completely separate from the test vignettes we used for the experiments outlined below.

\section{Methods}
\label{sec:methods}

We compared the accuracy and safety of the Babylon Triage and Diagnostic System against that of human doctors. Accuracy was assessed in terms of the relevance of the suggested conditions, and the appropriateness of the recommended triage action. Triage safety was assessed in terms of whether the suggested triage action was deemed safe (even if it was overly cautious).

\subsection{Outline}

The evaluation was performed using a semi-naturalistic role-play scenario that involved mock consultations between a patient and either a human doctor or the chatbot, based on realistic clinical vignettes. The role of doctors was played by general practitioners (GPs) who were hired on a locum basis for the experiment and who were not involved with the development of the model. Patients were played by GPs, some of whom were employees of Babylon, but none of whom were involved with the development of the model. We opted to use GPs to play the patients instead of professional actors as in a previous study \citep{middleton2016sorting}, to prioritise the accuracy of interpreting the information on the vignette over the realism of a layperson. One hundred clinical vignettes were created by independent medical practitioners who were not involved in the role-play experiment. Each vignette was designed to simulate a medical condition from the list of all conditions currently modelled by the Triage and Diagnostic System\footnote{
The list of conditions modelled by the Triage and Diagnostic System includes the majority of those encountered in General Practice in the United Kingdom, but does not include skin conditions, pregnancy-related conditions or paediatric conditions.
}, in a patient of at least 16 years of age. The vignettes contained information about the patient, their initial complaint(s), information about their symptoms and past medical history that should be offered on open questioning, and information that should only be reported on direct questioning. An example can be found in Supplementary Figure~\ref{fig:example_vignette}.

\subsection{Testing paradigm}
The study was conducted in four rounds over consecutive days. In each round, there were up to four ``patients'' and four doctors. Each patient was assigned a vignette as their presenting history and had independent consultations with each doctor and the Babylon Triage and Diagnostic System. This protocol was designed in the OSCE format to assess both history taking and diagnostic and triage accuracy. After each consultation the differential diagnosis and recommended triage produced by the doctor or Triage and Diagnostic System was recorded. In order to maintain blinding in the judging process, doctors selected their differential diagnoses from a list of all conditions covered by the Triage and Diagnostic System. Once the patient had completed consultations with all doctors and the Triage and Diagnostic System, they were assigned a new vignette and the process was repeated.

\section{Results}
\label{sec:results}

\begin{table*}[h]
\caption{Diagnostic performance for all seven doctors and the Babylon Triage and Diagnostic System (Babylon AI), in terms of the recall (sensitivity), precision (positive predictive value) and F1 score (harmonic mean of precision and recall) against the disease modelled by the clinical vignette. The diagnostic performance of the Babylon Triage and Diagnostic System is comparable to that of doctors.
\label{tab:doctor_performance}}
\centering
\begin{tabular}{>{\bf}l*{3}{>{$}c<{\%$}}>{$}c<{$}}
\toprule
& \multicolumn{1}{c}{\bf Average Recall} & %
\multicolumn{1}{c}{\bf Average Precision} & %
\multicolumn{1}{c}{\bf F1-score} & %
\multicolumn{1}{c}{\bf Number of Vignettes} \\ %
\midrule
Doctor A & 80.9  &  42.9   &   56.1   &   47   \\
Doctor B   &   64.1   &   36.8   &   46.7   &   78   \\
Doctor C   &   93.8   &   53.5   &   68.1   &   48   \\
Doctor D   &   84.3   &   38.1   &   52.5   &   51   \\
Doctor E   &   90.0   &   33.9   &   49.2   &   70   \\
Doctor F   &   90.2   &   43.3   &   58.5   &   51   \\
Doctor G   &   84.3   &   56.5   &   67.7   &   51   \\
\addlinespace[1ex]
Doctor Average    &   \textbf{83.9}   &   43.6   &   57.0   &   56.6   \\
Babylon AI &   80.0   &   \textbf{44.4}   &   \textbf{57.1}   &   100  \\
\bottomrule
\end{tabular}
\end{table*}

\subsection{Accuracy of differential diagnosis against vignette modelled disease}

We assessed the precision (also called positive predictive value) and recall (also called sensitivity) of the Babylon Triage and Diagnostic System and doctors against the condition modelled by the vignette. Recall is the proportion of relevant diseases that are included in the differential. When considering only the single disease modelled by the vignette, this corresponds to the proportion of differentials that contained the modelled disease, over all vignettes. Precision is the proportion of the diseases in the differential that are relevant. A precision of one hundred percent would be achieved if the differential diagnosis contained only the disease modelled by the vignette. In general this level of certainty is unlikely and even undesirable, given only the information provided on the vignette (i.e. in the absence of diagnostic tests), but penalises overly long differentials that would result in a higher recall.

In this study, the Babylon Triage and Diagnostic System was able to produce differential diagnoses with precision and recall comparable to that of doctors, and in some cases exceeded human level performance (Table~\ref{tab:doctor_performance}). The average recall of doctors was found to be 83.9\%, (64.1--93.8\%), meaning that doctors failed to include the vignette modelled disease in their differential in sixteen percent of cases on average.

\begin{figure*}
\centering
\includegraphics[width=\textwidth]{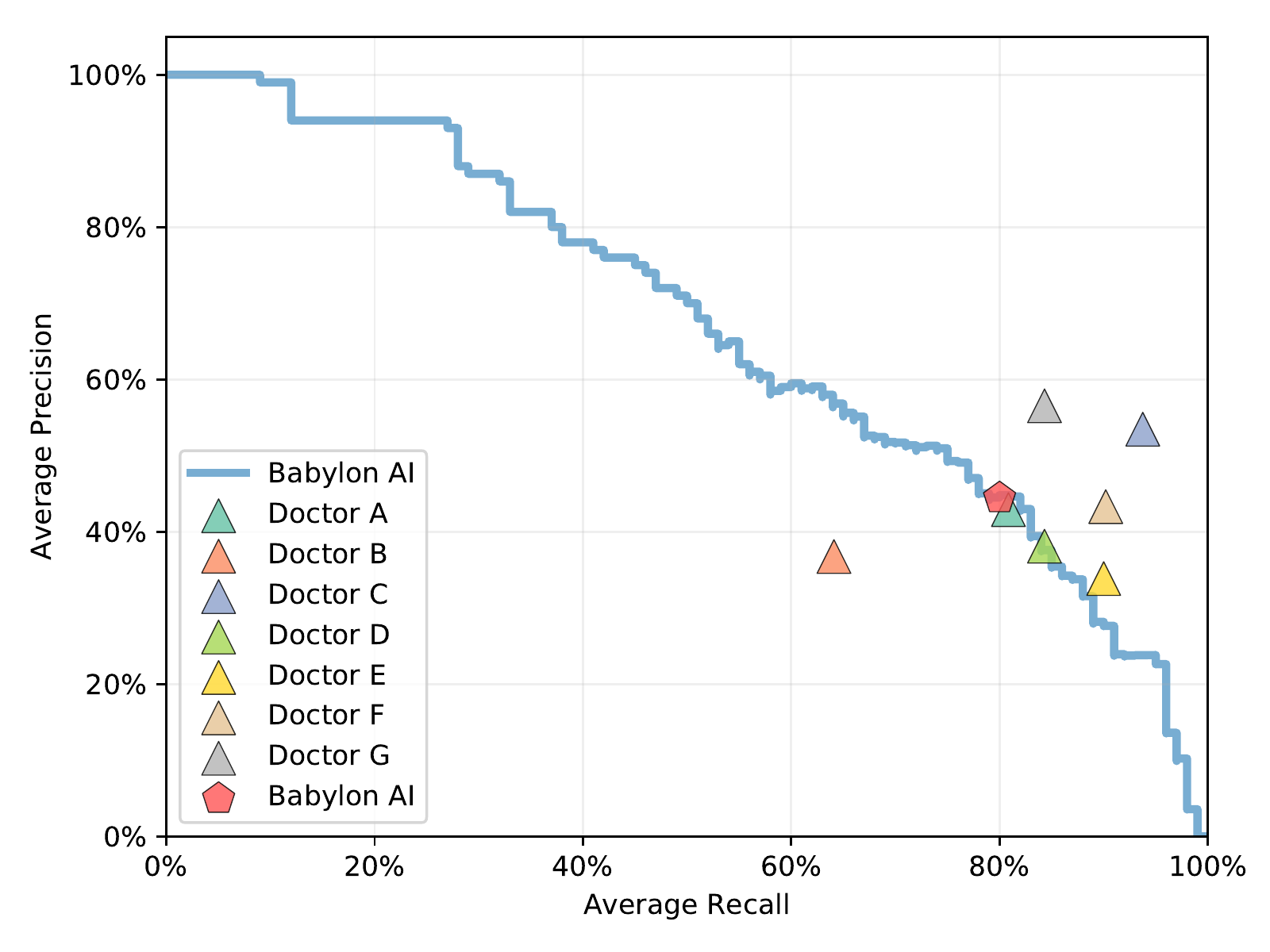}
\caption{Average recall and precision for doctors and for the Babylon Triage and Diagnostic System (Babylon AI) for different threshold parameters. Varying the internal thresholds allows the model to behave more similarly to different individual doctors, while maintaining a high level of performance, suggesting that it is not overly tuned to a particular operating point.}
\label{fig:main_PR_curve}
\end{figure*}

The Babylon Symptom Selector is based on a Bayesian model, which can calculate the posterior probabilities of all conditions in the model given the evidence known about a patient. Whether particular conditions are displayed to the user depends on whether they meet internal thresholds, defined by a combination of the probability and severity of these conditions. The threshold parameters used in the model are selected based on independent training vignettes but may be varied to allow a trade-off to be made between recall and precision. It is interesting to observe that different parameters can move the model's result closer to those of different doctors, for example towards Doctor D or E  (Figure~\ref{fig:main_PR_curve}), perhaps emulating the variability in individual doctors' preference for shorter, more precise differentials or longer, more exhaustive ones.

\subsection{Expert rating of differential diagnoses}

In addition to assessing the precision and recall compared to the disease modelled by the vignette, we also evaluated the overall differential diagnosis qualitatively. This was based on the intuition that, to be useful, a differential diagnosis must not only include the relevant diseases but also exclude diseases that are of little relevance to the patient's symptoms. To this end, we asked a senior medical practitioner\footnote{
Dr Benjamin A. White.
}, who was not part of the role play experiment, to serve as a judge and to rate the quality of the differentials produced both by the Babylon Triage and Diagnostic System and by doctors. The judge first reviewed the vignette and then rated all the differentials for this vignette on a four point scale \textit{(poor, okay, good, excellent)}. A differential was rated ``excellent'' if the judge could not find any issues with it, ``good'' if it had minor issues (such as the omission of a slightly irrelevant conditions, or if the order of the differential was deemed imperfect), ``okay'' if the list of conditions was generally acceptable, and ``poor'' if it was unacceptable (such as the omission of the most important conditions, or the inclusion of diseases completely unrelated to the presenting symptoms). The differentials were shown in random order and the judge was blinded to whether the differential had been produced by a human or the Babylon Triage and Diagnostic System. We also repeated this process with two in-house GPs. 

\begin{figure}
\begin{center}
\includegraphics[width=\columnwidth]{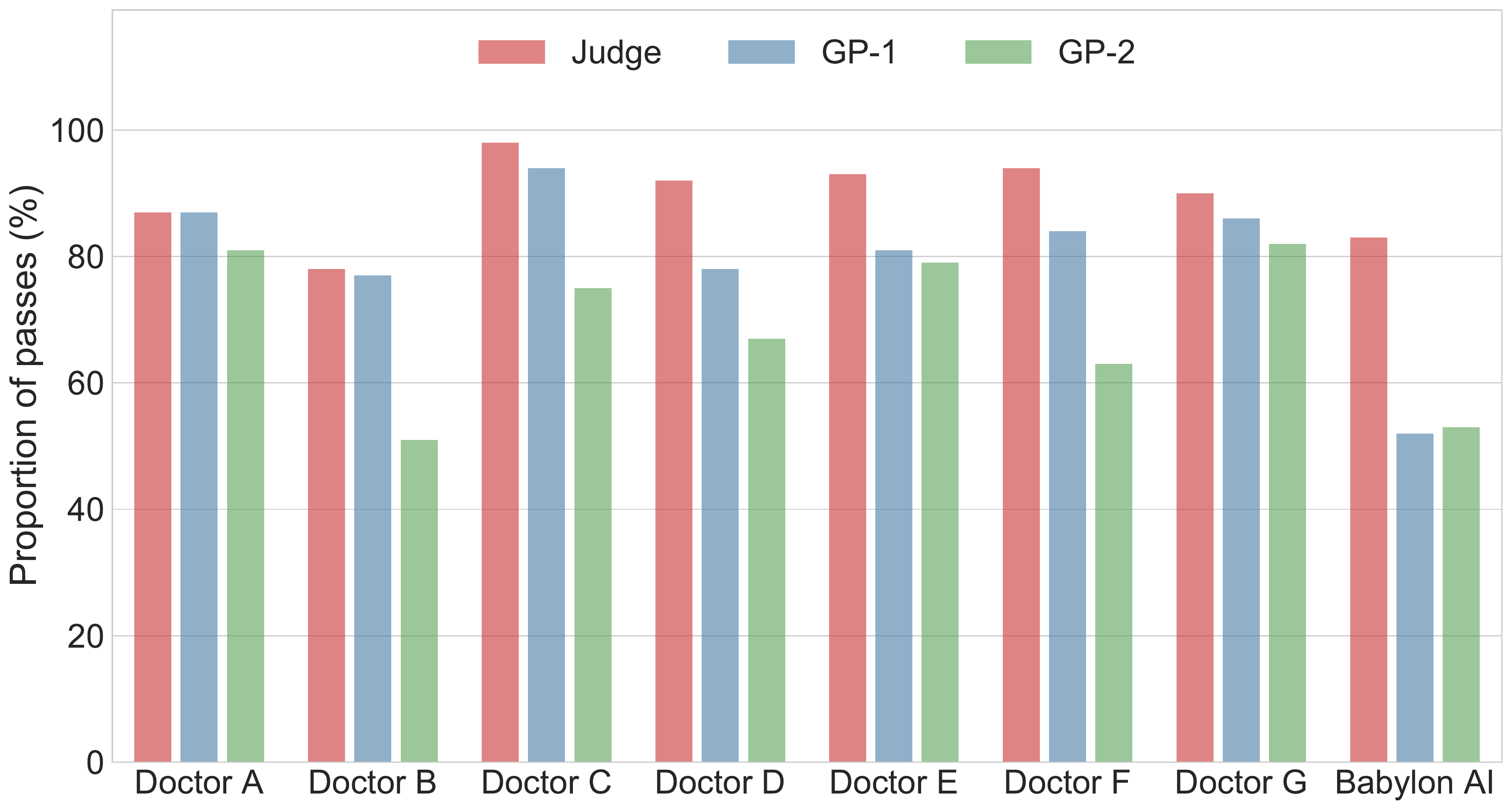}
\caption{Percentage of differential diagnoses rated as "okay" or better by the judge and the two GPs for doctors and the Babylon Triage and Diagnostic System (Babylon AI). There is considerable disagreement between the three ratings, suggesting the qualitative assessment of differential diagnoses might be influenced by personal preference.}\label{fig:fig_all_judges}
\end{center}
\end{figure}

We found that there was considerable disagreement between the medical practitioners' subjective assessment of the differentials (see Figure~\ref{fig:fig_all_judges}; Supplementary Tables~\ref{tab:differential_diag_perf},~\ref{tab:differential_diag_perf_gp1}~and~\ref{tab:differential_diag_perf_gp2}). For the judge, the lists of diseases output by the Babylon Triage and Diagnostic System were found to be of comparable quality to those produced by doctors (83.0\% rated ``okay'' or better, compared to 78.2--97.9\%). The same was the case for one of the GPs (GP-2), who was generally harsher on the evaluation (53.0\% rated ``okay'' or better, compared to 51.3--82.4\%). However, another GP (GP-1) rated the quality of differentials of the Babylon Triage and Diagnostic System lower than those of doctors (52.0\% rated ``okay'' or better, compared to 76.9--93.8\%).

\begin{figure}
\begin{center}
\includegraphics[width=\columnwidth]{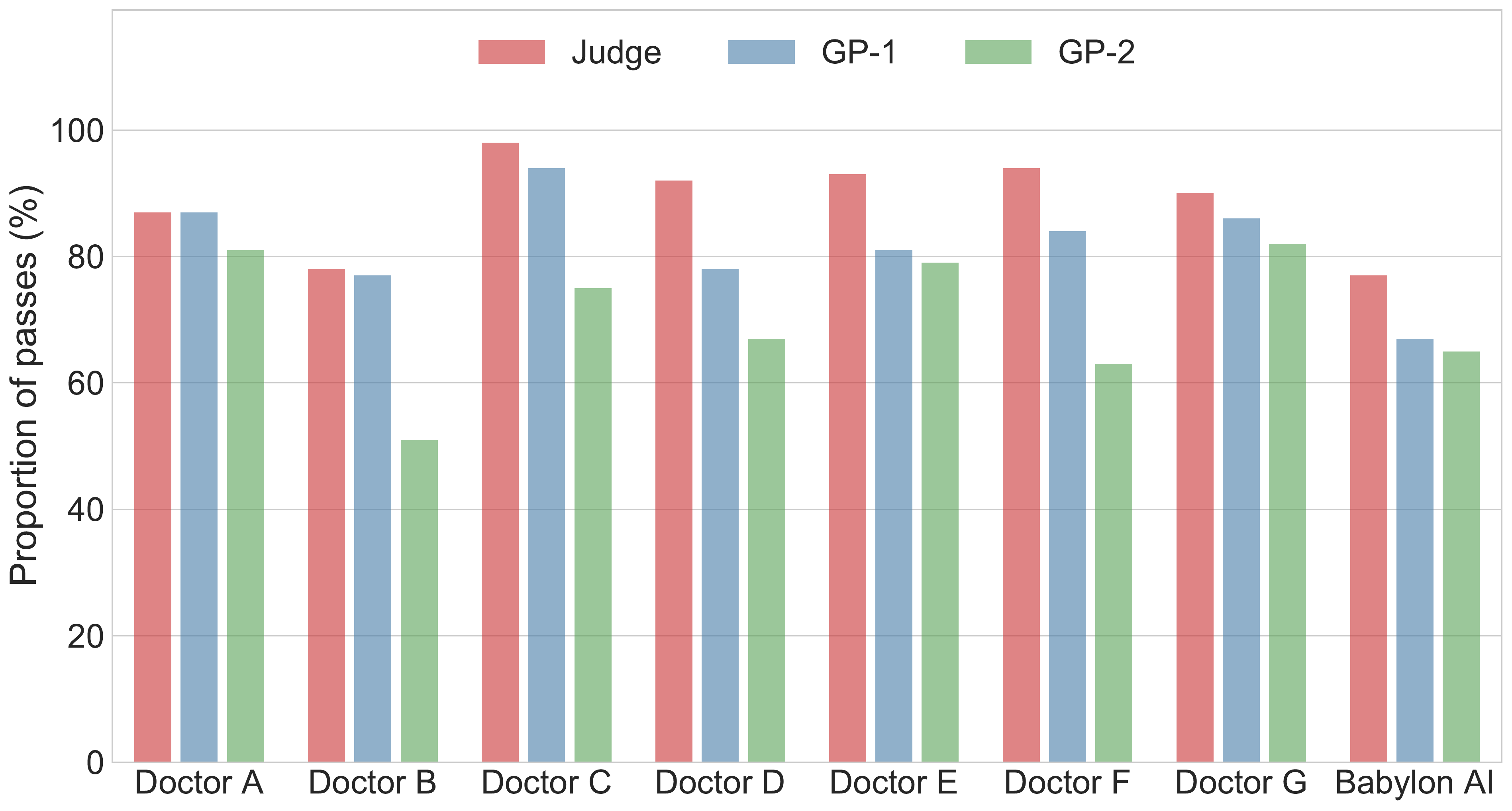}
\caption{Percentage of differential diagnoses rated as "okay" or better by the judge and the two GPs for doctors and the Babylon Triage and Diagnostic System (Babylon AI), where the latter is tuned to provide higher precision (at the expense of recall). The differentials provided by the Babylon Triage and Diagnostic System were rated to be of comparable quality to those of  doctors.}\label{fig:fig_all_judges_new_thresholds}
\end{center}
\end{figure}

We considered that the disparity in the qualitative evaluation of differential diagnoses might be due to a difference in interpretation and that some medical practitioners might be less tolerant of disease lists that are long or contain less relevant diseases, even if the relevant conditions are included. Although we don't have sufficient statistical power to test this hypothesis, we repeated the experiment with the Babylon Triage and Diagnostic System tuned to provide higher precision at the expense of lower recall (Supplementary Figure~\ref{fig:main_PR_curve}). This mode resulted in a marked improvement in the ratings of the GPs, which anecdotally suggests a preference for more concise differentials for these individuals (Figure~\ref{fig:fig_all_judges_new_thresholds}).

\subsection{Assessment of triage safety and appropriateness}

\begin{table}[h]
\caption{Safety and appropriateness of triage recommendations for doctors and the Babylon Triage and Diagnostic System (Babylon AI) against a range of acceptable recommendations provided by an independent judge. The Babylon Triage and Diagnostic System gives safer triage recommendations than the doctors on average, at the expense of a marginally lower appropriateness.
\label{tab:triage_safety}}
\centering
\begin{tabular}{>{\bf}l*{2}{>{$}c<{\%$}}>{$}c<{$}}
\toprule
& \multicolumn{1}{c}{\bf Safety} & \multicolumn{1}{c}{\bf Appr.} & \multicolumn{1}{c}{\bf Cases} \\
\midrule
Doctor A & 95.7 & 91.5 & 47 \\ 
Doctor B & 89.7 & 89.7 & 78 \\
Doctor C & 100.0 & 93.8 & 48 \\
Doctor D & 94.1 & 94.1 & 51 \\
Doctor E & 90.0 & 85.7 & 70 \\
Doctor F & 94.1 & 90.2 & 51 \\
Doctor G & 88.2 & 88.2 & 51 \\
\addlinespace[1ex]
Doctor Average & 93.1 & \bf 90.5 & 56.6 \\
Babylon AI & \bf 97.0 & 90.0 & 100 \\
\bottomrule

\end{tabular}
\end{table}

\begin{figure*}[htb]
\centering
\includegraphics[width=\textwidth]{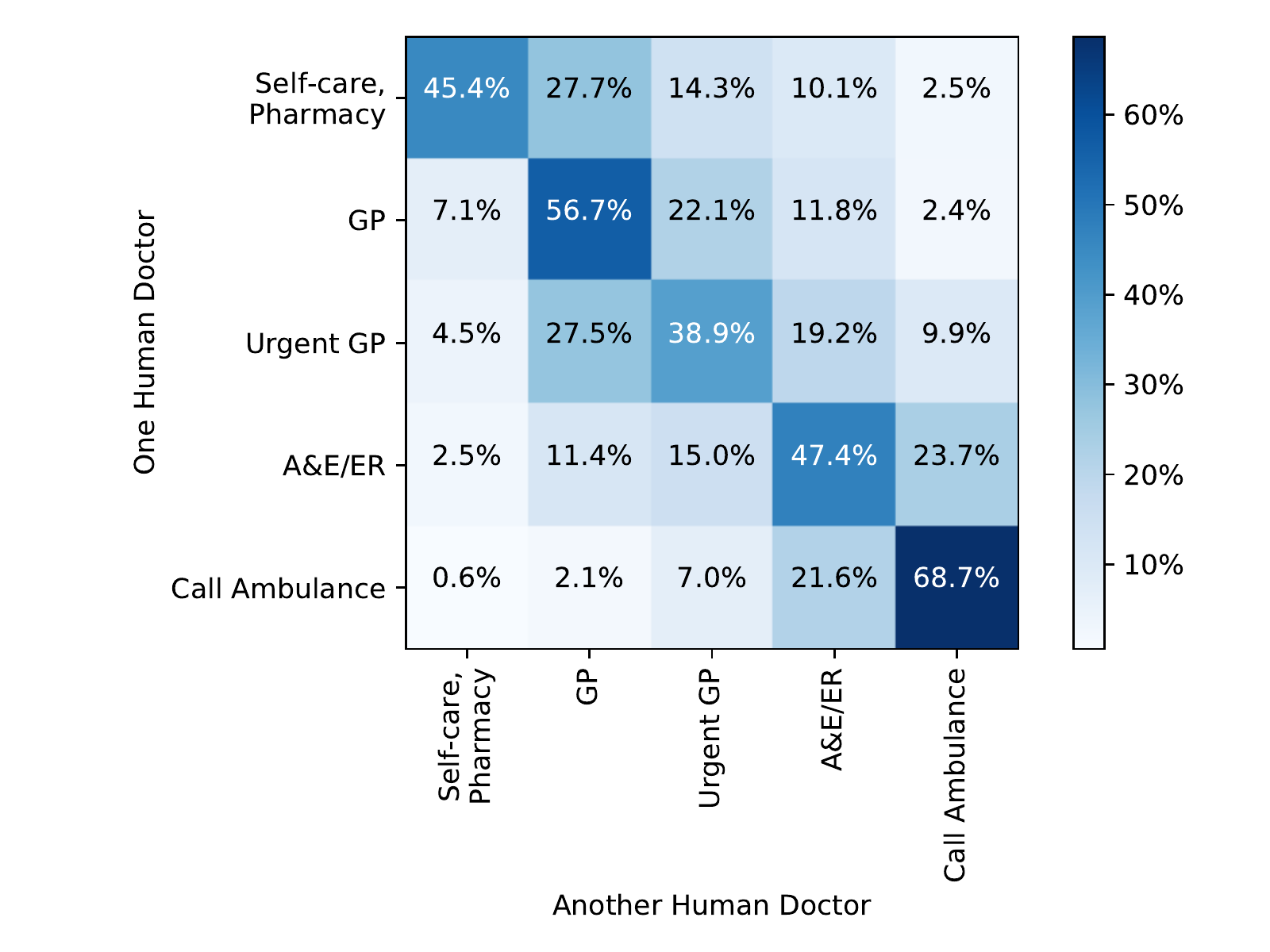}
\caption{Confusion matrix between a single human doctor another human doctor (aggregated over all pairwise combinations of doctors). Considerable disagreement exists between the triage recommendations of different doctors, with confusion between all pairs of triage categories. Note that the {\it self-care} and {\it pharmacy} categories have been combined.
\label{fig:doctor_vs_doctor_confusion}}
\end{figure*}

In addition to rating the quality of doctors differential diagnoses, the expert judge was also asked  to specify a range of safe and appropriate triage outcomes for each vignette. Providing a range of acceptable triage recommendations was motivated by the fact that doctors often disagree on the most appropriate triage recommendation (Supplementary Figure~\ref{fig:doctor_vs_doctor_confusion}), however it is not necessarily the case that any of these different opinions are inappropriate or unsafe \citep{cathain2003}. By providing the minimum and maximum appropriate triage, our judge indicates the range of recommendations that are neither unsafe nor overly cautious. 

We compared the triage recommendations of doctors and the Babylon Triage and Diagnostic System against the judge's ``gold standard'' range. We define a ``safe'' triage as any recommendation which was of equal or greater urgency than the judge's minimum triage, and an ``appropriate'' triage as any recommendation that fell within the judge's range of acceptable recommendations. In this study, we found that the Babylon Triage and Diagnostic System provided a safer triage recommendation than doctors on average (97.0\% versus 93.1\%), at the expense of a marginally lower appropriateness (90.0\% versus 90.5\%; see Table~\ref{tab:triage_safety}).

We repeated this process with three in-house GPs and found the triage safety and appropriateness of the Babylon Triage and Diagnostic System relative to the doctors to be consistent with those of the judge, although the scores from the GPs were found to be lower for both the Babylon Triage and Diagnostic System and the doctors (Table~\ref{tab:triage_recommendations}).

\begin{table*}[h]
\caption{Safety and appropriateness of triage recommendations for doctors and the Babylon Triage and Diagnostic System (Babylon AI) against a range of acceptable recommendations provided by GPs. The AI powered System gives safer triage recommendations than the doctors on average, at the expense of a slightly lower appropriateness.
\label{tab:triage_recommendations}}
\centering
\begin{tabular}{>{\bf}l*{6}{>{$}c<{\%$}}}
\toprule
& \multicolumn{2}{c}{\bf GP-1} %
& \multicolumn{2}{c}{\bf GP-2} %
& \multicolumn{2}{c}{\bf GP-3} \\ 
& \multicolumn{1}{c}{\bf Safety} %
& \multicolumn{1}{c}{\bf Appr.}  %
& \multicolumn{1}{c}{\bf Safety} %
& \multicolumn{1}{c}{\bf Appr.}  %
& \multicolumn{1}{c}{\bf Safety} %
& \multicolumn{1}{c}{\bf Appr.} \\ %
\midrule
Doctor A & 97.9 & 89.4 & 91.5 & 83.0 & 95.7 & 89.4 \\ 
Doctor B & 79.5 & 75.6 & 60.3 & 59.0 & 75.6 & 74.4 \\ 
Doctor C & 97.9 & 89.6 & 93.8 & 89.6 & 95.8 & 93.8 \\ 
Doctor D & 80.4 & 76.5 & 64.7 & 62.8 & 86.3 & 84.3 \\ 
Doctor E & 84.3 & 78.6 & 70.0 & 67.1 & 80.0 & 78.6 \\ 
Doctor F & 92.2 & 86.3 & 74.5 & 68.6 & 92.2 & 84.3 \\ 
Doctor G & 92.2 & 88.2 & 72.6 & 70.6 & 84.3 & 80.4 \\
\addlinespace[1ex]
Doctor Average & 89.2 & \bf 83.5 & 75.3 & 71.5 & 87.1 & \bf 83.6 \\
Babylon AI  & \bf 90.0 & 74.0 & \bf 81.0 & \bf 75.0 & \bf 90.0 & 81.0 \\
\bottomrule
\end{tabular}
\end{table*}

\subsection{Performance against publicly available case vignettes}

In order to provide a benchmark against previous work, as well as to the diagnostic accuracy that is expected for human practitioners, we assessed the performance of the Babylon Triage and Diagnostic System against three sets of publicly available case vignettes. These were case vignettes used in a previous study \citep{semigran2015evaluation}, and from preparation materials for the MRCGP Clinical Skills Assessment (CSA) and Applied Knowledge Test (AKT), focusing on the diagnostic component of the curriculum (in contrast to content assessing management).

\subsubsection{\citealt{semigran2015evaluation} Vignettes}

The methodology described previously was repeated for 30 vignettes from a previous study by \citep{semigran2015evaluation}. We excluded vignettes from the original study that were based on conditions that are outside of the scope of the Babylon Triage and Diagnostic System, consistent with the original methodology. Specifically, these included paediatric and dermatological conditions, and tetanus which is not currently in the model yet based on its very low incidence rate in the United Kingdom. These vignettes were tested against both the Babylon Triage and Diagnostic System and three doctors. As per the original study, we report the recall of the condition modelled by the vignette for the top 1 and top 3 conditions listed in the differential. The Babylon Triage and Diagnostic System identified the modelled condition as its top 1 in 21 out of 30 vignettes (70.0\%) and in its top 3 in 29 out of 30 vignettes (96.7\%). On average, doctors identified the modelled condition in their top 1 in 75.3\% of vignettes and in their top 3 in 90.3\% of vignettes. This demonstrates a significant improvement relative to other symptom checkers evaluated in the original study.

\subsubsection{MRCGP Applied Knowledge Test (AKT)}

The Babylon Triage and Diagnostic System was tested against 15 AKT vignettes (based on \citealt{RCGPLearning}) by providing all available information about the patient to the model, consistent with how an AKT exam participant would see the exam question (since there is no history-taking component). The correct condition appeared in the top 3 conditions suggested by the Babylon Triage and Diagnostic System in 13 of 15 vignettes (86.7\%).

\subsubsection{MRCGP Clinical Skills Assessment (CSA)}

The Babylon Triage and Diagnostic System was tested against 36 CSA vignettes (translated from \citealt{Wadhera2011}; and \citealt{Stannett2016}) by first providing the main presenting complaint, then simulating a user's interaction with the Babylon Triage and Diagnostic System by only providing information to the model if the Triage and Diagnostic System would directly ask about it. This is consistent with the CSA exam which requires the participant to take a medical history. The modelled disease appeared in the top 3 conditions suggested by the Triage and Diagnostic System in 27 of 36 vignettes (75.0\%).

\section{Discussion}
\label{sec:discussion}

We performed a prospective validation study of the accuracy and safety of an AI powered Triage and Diagnostic System, using an experimental paradigm designed to simulate realistic consultations. Overall we found that the Babylon Triage and Diagnostic System was able to identify the condition modelled by a clinical vignette with accuracy comparable to human doctors (in terms of precision and recall). We also found that the triage advice recommended by the Babylon Triage and Diagnostic System was safer on average than human doctors, when compared to the ranges provided by independent expert judges, with only minimal reduction in appropriateness. In other words, the AI system was able to safely triage patients without reverting to overly pessimistic fallback decisions. 

\begin{figure*}[htb]
\centering
\includegraphics[width=\textwidth]{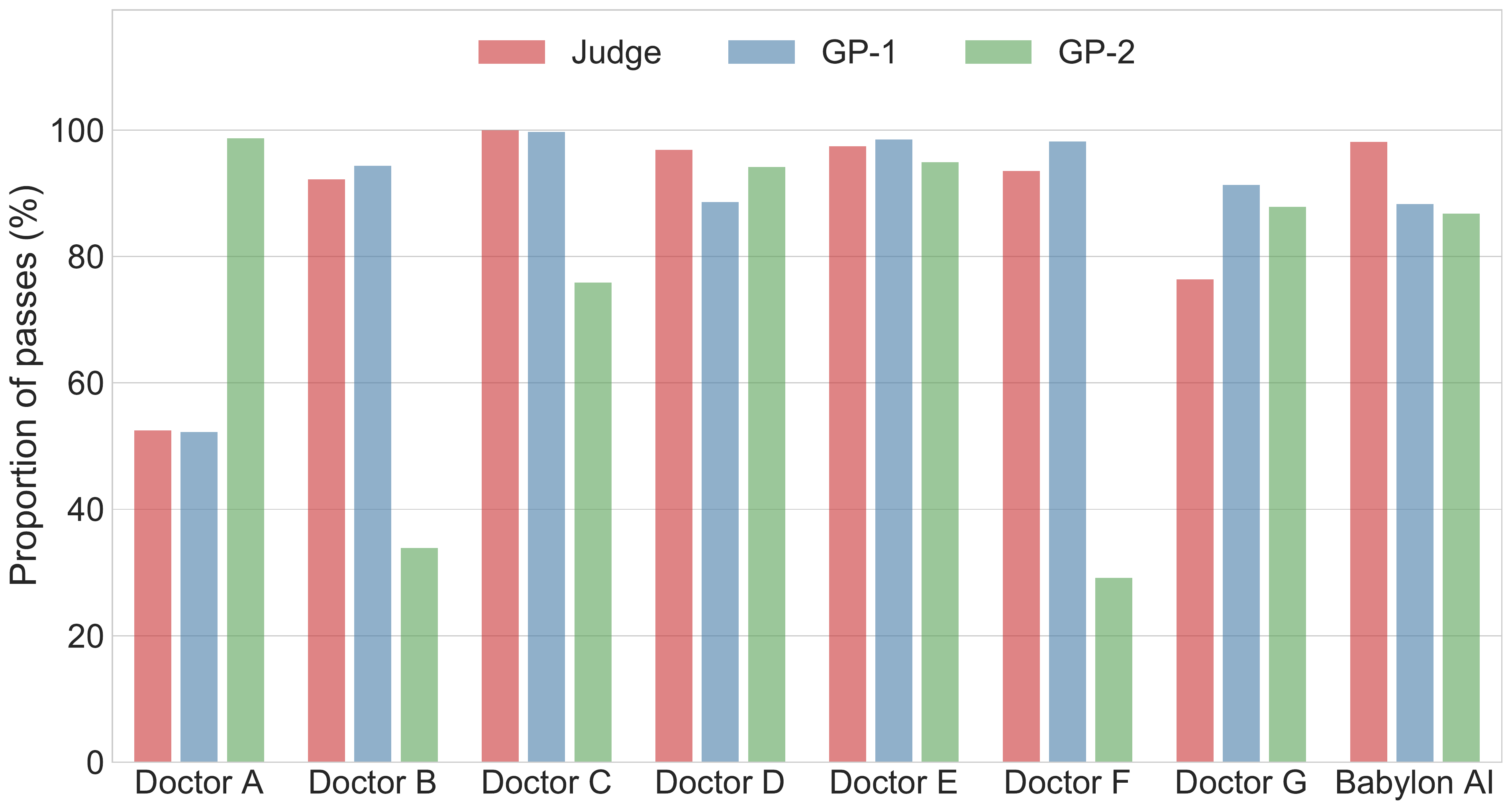}
\caption{Percentage of differential diagnoses rated as ``okay'' or better by the judge and the two GPs for human doctors and the Babylon Triage and Diagnostic System (Babylon AI), after reweighting by the annual incidence of the disease modelled by the vignette.
\label{fig:all_judges_modelled_diseases}}
\end{figure*}

We adopted a test protocol using simulated clinical vignettes which allowed us to evaluate a combination of common and rare conditions, the latter of which would be difficult to evaluate without a clinical trial with a sample size large enough to contain diseases with low incidence rates. While this might be considered a strength of our study, since it is not biased towards common presentations, our results cannot be directly interpreted with respect to real-world accuracy and safety. To illustrate the differences that might be expected in a real-world study, we reweighted our results by the annual incidence of the modelled disease for each vignette. We found that the accuracy and rating of differentials produced by the Babylon Triage and Diagnostic System improved compared to those of doctors after accounting for disease incidence (Supplementary Table~\ref{tab:modelled_disease_perf} and Supplementary Figure~\ref{fig:all_judges_modelled_diseases}). Surprisingly, we found that the accuracy and rating of some doctors decreased considerably after reweighting. This is likely due to the fact that the most common conditions carry substantially more weight than the rarer ones; thus the results will be highly sensitive to a few vignettes (in particular, Doctor A did not include a modelled disease in their differential for a vignette, where that modelled disease was very common and hence had high weight). Further work will be required to more rigorously investigate the diagnostic accuracy in a real-world clinical setting.

One source of bias in this study derives from the limitation imposed on doctors to only select diseases that are modelled in the Babylon Triage and Diagnostic System. As the "correct" disease for each vignette was always from this list, this may have provided human doctors with some advantage in terms of precision and recall compared to free text entry. However, it would have also constrained them from providing a fuller and more nuanced differential diagnosis overall, which may have disadvantaged them in terms of judge rating of overall differential quality. The intention in assigning this limitation as part of the testing protocol was to ensure blinding when the judges assessed the quality of the differential diagnosis.

The Babylon Triage and Diagnostic System listed the correct disease among the top 3 differentials in 86.7\% of AKT cases and 75.0\% of CSA cases, for the limited question set that was tested. However, it is important to note that the components of the MRCGP examination used to assess the Babylon Triage and Diagnostic System were limited to those based on history taking and diagnostics. The full MRCGP examination (a combination of the AKT and CSA) also assesses a wide range of other skills not examined in this study. Therefore, the performance of the Babylon Triage and Diagnostic System in this study cannot be taken as a demonstration of an ability to pass the examination in full, but on the diagnostic component in isolation, it achieved accuracy rates above 72\%, the average pass mark for the past 5 years \citep{RCGP2018one,RCGP2018two} for the entire MRCGP exam.

Another possible limitation of our study is that we evaluated only clinical cases that were based on a single underlying condition (although we did include past medical history and pre-existing conditions). In reality, patients may have multiple undiagnosed diseases. However, one of the strengths of our approach, which uses a Bayesian model, is that it is able to reason about multiple causes of a patient's presenting symptoms. It would be useful to test whether the performance relative to doctors is different in cases where multiple diseases must be diagnosed.

Finally, this study emphasises the difficulty in objectively evaluating the accuracy of a differential diagnosis. Even when the true underlying condition is identified, the quality of the overall differential may be poor due to the omission of important alternative explanations for a patient's symptoms, or the inclusion of irrelevant diseases. By evaluating differential diagnoses qualitatively using independent judges, we found that considerable disagreement exists in the subjective rating by different individuals, including differential diagnoses of human doctors. This may be due to the fact that a judge's rating is itself based on personal assessment of the clinical case, which may be prone to error, or due to differences in personal preference for longer or shorter differential diagnoses. Ultimately, there is likely no adequate ``gold standard'' differential diagnosis, and future work would benefit from assessing the inter-rater agreement between a larger sample of doctors.

\section{Conclusion}
\label{sec:conclusion}

Artificial intelligence powered symptom checkers have the potential to provide diagnostic and triage advice with a level of accuracy and safety approaching that of human doctors. Such systems may hold the promise of reduced costs and improved access to healthcare worldwide, but realising this requires greater levels of confidence from the medical community and the wider public. Key to this confidence is a better understanding of the strengths and weaknesses of human doctors, who do not always agree on the cause of a patient's symptoms or the most appropriate triage outcome, and an improved awareness of the accuracy and safety of AI powered systems. Further studies using larger, real-world cohorts will be required to demonstrate the relative performance of these systems to human doctors.

\section{Acknowledgements}
\label{sec:acknowledgements}

Independent medical practitioners helped us create the testing vignettes for the main experiment of our paper. They were: Prof. Megan Mahoney from Stanford University, Dr Benjamin A. White from Harvard Medical School/Massachusetts General Hospital, Department of Emergency Medicine. The authors would like to thank Dr Benjamin A. White, Dr Hannah Allen, Dr Marieke Reddingius for their help rating the differentials and triage.

\newpage

\bibliographystyle{plainnat}
\bibliography{references}

\beginsupplement
\clearpage
\onecolumn
\noindent\large\textbf{SUPPLEMENTARY MATERIAL}

\begin{table*}[htb]
\caption{Independent assessment of the quality of differential diagnosis by the judge. Each differential from the human doctors and the Babylon Triage and Diagnostic System (Babylon AI) was rated on a four point scale (poor/okay/good/excellent). The subjective quality of the Babylon Triage and Diagnostic System was found to be within the ``Pass'' range of values for human doctors.
\label{tab:differential_diag_perf}}
\centering
\begin{tabular}{>{\bf}l*{5}{>{$}c<{\%$}}>{$}c<{$}}
\toprule
& \multicolumn{1}{c}{\bf Poor} & 
\multicolumn{1}{c}{\bf Okay} & 
\multicolumn{1}{c}{\bf Good} & 
\multicolumn{1}{c}{\bf Excellent} & 
\multicolumn{1}{c}{\bf Pass} & 
\multicolumn{1}{c}{\bf Cases} \\ 
\midrule
Doctor A & 12.8 & 25.5 & 55.3 & 6.4 & 87.2 & 47 \\ 
Doctor B & 21.8 & 33.3 & 37.2 & 7.7 & 78.2 & 78 \\ 
Doctor C & 2.1 & 41.7 & 47.9 & 8.3 & 97.9 & 48 \\ 
Doctor D & 7.8 & 17.7 & 62.8 & 11.8 & 92.2 & 51 \\ 
Doctor E & 7.1 & 5.7 & 65.7 & 21.4 & 92.9 & 70 \\ 
Doctor F & 5.9 & 15.7 & 74.5 & 3.9 & 94.1 & 51 \\ 
Doctor G & 9.8 & 43.1 & 43.1 & 3.9 & 90.2 & 51 \\ 
\addlinespace[1ex]
Doctor Average & 9.6 & 26.1 & 55.2 & 9.1 & 90.4 & 56.6 \\
Babylon AI  & 17.0 & 34.0 & 44.0 & 5.0 & 83.0 & 100 \\
\bottomrule
\end{tabular}
\end{table*}

\begin{table*}[htb]
\caption{Independent assessment of the quality of differential diagnosis by GP-1. The subjective quality of the Babylon Triage and Diagnostic System (Babylon AI) was found to be out of the ``Pass'' range of values for human doctors by this GP.
\label{tab:differential_diag_perf_gp1}}
\centering
\begin{tabular}{>{\bf}l*{5}{>{$}c<{\%$}}>{$}c<{$}}
\toprule
& \multicolumn{1}{c}{\bf Poor} & 
\multicolumn{1}{c}{\bf Okay} & 
\multicolumn{1}{c}{\bf Good} & 
\multicolumn{1}{c}{\bf Excellent} & 
\multicolumn{1}{c}{\bf Pass} & 
\multicolumn{1}{c}{\bf Cases} \\ 
\midrule
Doctor A&12.8 &31.9 &34.0 &21.3 &87.2 &47\\
Doctor B&23.1 &28.2 &28.2 &20.5 &76.9 &78\\
Doctor C&6.3&27.1 &47.9 &18.8 &93.8 &48\\
Doctor D&21.6 &23.5 &35.3 &19.6 &78.4 &51\\
Doctor E&18.6 &21.4 &42.9 &17.1 &81.4 &70\\
Doctor F&15.7 &23.5 &41.2 &19.6 &84.3 &51\\
Doctor G&13.7 &15.7 &41.2 &29.4 &86.3 &51\\
\addlinespace[1ex]
Doctor Average&16.0 &24.5 &38.7 &20.9 &84.0 &56.6\\
Babylon AI &48.0 &13.0 &27.0 &12.0 &52.0 &100\\
\bottomrule
\end{tabular}
\end{table*}

\begin{table*}[htb]
\caption{Independent assessment of the quality of differential diagnosis by GP-2. Each differential from the human doctors and the Babylon Triage and Diagnostic System (Babylon AI) was rated on a four point scale (poor/okay/good/excellent). The subjective quality of the Babylon Triage and Diagnostic System was found to be within the ``Pass'' range of values for human doctors, similar to the judge's results.
\label{tab:differential_diag_perf_gp2}}
\centering
\begin{tabular}{>{\bf}l*{5}{>{$}c<{\%$}}>{$}c<{$}}
\toprule
& \multicolumn{1}{c}{\bf Poor} & 
\multicolumn{1}{c}{\bf Okay} & 
\multicolumn{1}{c}{\bf Good} & 
\multicolumn{1}{c}{\bf Excellent} & 
\multicolumn{1}{c}{\bf Pass} & 
\multicolumn{1}{c}{\bf Cases} \\ 
\midrule
Doctor A&19.1 &29.8 &14.9 &36.2 &80.9 &47\\
Doctor B&48.7 &15.4 &9.0 &26.9 &51.3 &78\\
Doctor C&25.0 &16.7 &22.9 &35.4 &75.0 &48\\
Doctor D&33.3 &15.7 &23.5 &27.5 &66.7 &51\\
Doctor E&21.4 &20.0 &28.6 &30.0 &78.6 &70\\
Doctor F&37.2 &13.7 &21.6 &27.5 &62.8 &51\\
Doctor G&17.7 &11.8 &27.5 &43.1 &82.4 &51\\
\addlinespace[1ex]
Doctor Average&28.9 &17.6 &21.1 &32.4 &71.1 &56.6\\
Babylon AI &47.0 &11.0 &6.0 &36.0 &53.0 &100\\
\bottomrule
\end{tabular}
\end{table*}

\begin{figure*}[htb]
\centering
\includegraphics[width=0.9\textwidth]{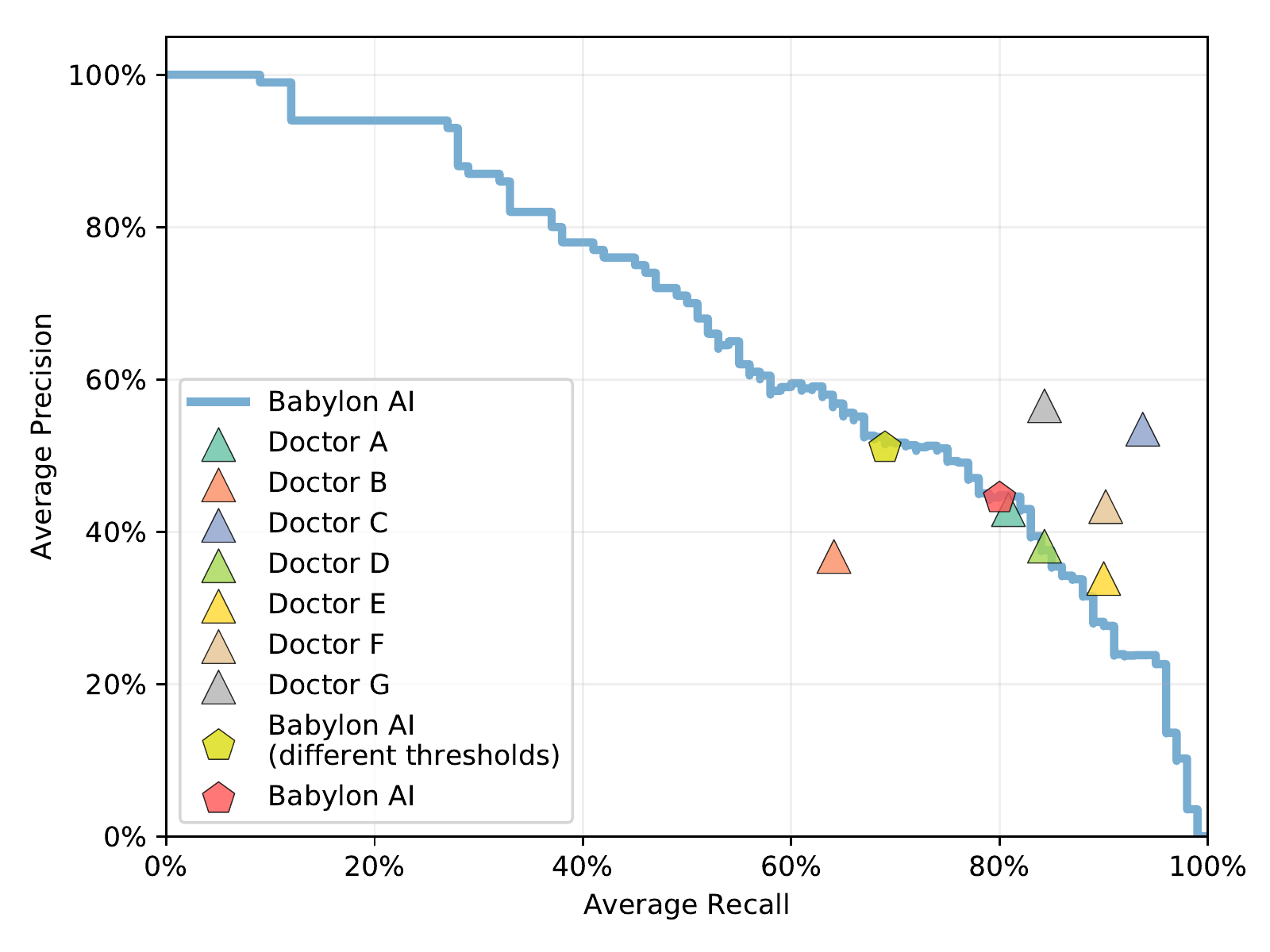}
\caption{Average recall and precision for the Babylon Triage and Diagnostic System (Babylon AI) for different internal threshold parameters, showing the operating point typically used (red pentagon) and an alternative operating point (yellow pentagon) that favours higher precision at the expense of reduced recall.}
\label{fig:PR_different_thresholds}
\end{figure*}

\begin{table*}[htb]
\caption{Diagnostic performance for all seven doctors and the Babylon Triage and Diagnostic System (Babylon AI), in terms of the recall (sensitivity), precision (positive predictive value) and F1 score (harmonic mean of precision and recall) against the disease modelled by the clinical vignette, after reweighting by the annual incidence of the disease modelled by the vignette.
\label{tab:modelled_disease_perf}}
\centering
\begin{tabular}{>{\bf}l*{3}{>{$}c<{\%$}}>{$}c<{$}}
\toprule
& \multicolumn{1}{c}{\bf Recall} & 
\multicolumn{1}{c}{\bf Av. Precision} & 
\multicolumn{1}{c}{\bf F1-score} & 
\multicolumn{1}{c}{\bf Cases} \\ 
\midrule
Doctor A&52.0 &24.6 &33.4 &47\\
Doctor B&86.5 &37.2 &52.0 &78\\
Doctor C&99.96 &47.0 &64.0 &48\\
Doctor D&94.0 &33.6 &49.5 &51\\
Doctor E&96.3 &39.4 &55.9 &70\\
Doctor F&93.1 &50.2 &65.2 &51\\
Doctor G&75.1 &56.7 &64.6 &51\\
\addlinespace[1ex]
Doctor Average&85.3 &41.2 &55.0 &56.6\\
Babylon AI&97.9 &83.3 &90.0 &100\\
\bottomrule
\end{tabular}
\end{table*}

\begin{figure*}[htb]
\centering
\begin{tcolorbox}[%
fonttitle=\bfseries,%
title=Vignette Example,%
colback=gray!5,%
width=.55\textwidth]
  {\bf Modelled Disease:}
  \begin{itemize}
  \item Allergic Rhinitis
  \end{itemize}
  
  {\bf Patient:}
  \begin{itemize}
  \item Male, Age 35
  \end{itemize}

  {\bf Presenting Complaint / Initial user input:}
  \begin{itemize}
  \item ``I've been sneezing and have a stuffy nose.''
  \end{itemize}

  {\bf History, on open questioning:}
  \begin{itemize}
  \item Sneezing and stuffy nose occur every spring
  \item Other symptoms include itchy eyes and runny nose.
  \item Can be worse after going outdoors.
  \end{itemize}

  {\bf If asked directly:}
  \begin{itemize}
  \item Occasional Cough.
  \item Itchy sensation in the throat
  \item Eyes can be red with a little mucus.
  \item Frequent throat clearing
  \item No fever, chills or difficulty breathing
  \item Feels better with Benadryl but it makes him sleepy.
  \item General malaise with the above symptoms.
  \end{itemize}

  {\bf Family History:}
  \begin{itemize}
  \item Father and mother are healthy
  \end{itemize}

  {\bf Past Medical History:}
  \begin{itemize}
  \item Asthma as a child
  \end{itemize}

  {\bf Allergies:}
  \begin{itemize}
  \item NKDA
  \end{itemize}
  
\end{tcolorbox}

\caption{An example of a test vignette.}
\label{fig:example_vignette}
\end{figure*}

\end{document}